\documentclass[
]{ceurart}

\usepackage{listings}
\usepackage{tabularx}
\usepackage{todonotes}

\begin{document}

\copyrightyear{2023}
\copyrightclause{Copyright for this paper by its authors.
  Use permitted under Creative Commons License Attribution 4.0
  International (CC BY 4.0).}

\conference{CLEF 2023: Conference and Labs of the Evaluation Forum, September 18–21, 2023, Thessaloniki, Greece}

\title{Text Simplification of Scientific Texts for Non-Expert Readers}

\title[mode=sub]{Notebook for the SimpleText Lab at CLEF 2023}

\author[1]{Björn Engelmann}[%
orcid=0009-0000-7074-9066,
email=bjoern.engelmann@th-koeln.de,
]

\author[1]{Fabian Haak}[%
orcid=0000-0002-3392-7860,
email=fabian.haak@th-koeln.de,
]

\author[1]{Christin Katharina Kreutz}[%
orcid=0000-0002-5075-7699,
email=christin.kreutz@th-koeln.de,
]

\author[1]{Narjes {Nikzad Khasmakhi}}[%
orcid=0000-0003-3536-1343,
email={narjes.nikzad_khasmakhi@th-koeln.de},
]

\author[1]{Philipp Schaer}[%
orcid=0000-0002-8817-4632,
email=philipp.schaer@th-koeln.de,
]

\address[1]{TH Köln – University of Applied Sciences, Cologne, Germany}


\begin{abstract}
%
%
%
Reading levels are highly individual and can depend on a text's language, a person's cognitive abilities, or knowledge on a topic. Text simplification is the task of rephrasing a text to better cater to the abilities of a specific target reader group.
Simplification of scientific abstracts helps non-experts to access the core information by bypassing formulations that require domain or expert knowledge.
This is especially relevant for, e.g., cancer patients reading about novel treatment options.

The SimpleText lab hosts the simplification of scientific abstracts for non-experts (Task 3) to advance this field.
We contribute three runs employing out-of-the-box summarization models (two based on T5, one based on PEGASUS) and one run using ChatGPT with complex phrase identification.

\end{abstract}


\maketitle

\section{Introduction}

Even with the ever-increasing number of scientific publications, accessing information from scientific texts for non-expert readers remains a problem, e.g., cancer patients investigating possible treatment options~\cite{DBLP:conf/flairs/AfsarCF21}.
Complex scientific texts pose challenges for non-expert readers for various reasons: jargon, abbreviations, and technical details hinder accessing such content.  
Text simplification attempts to mitigate these challenges, making the content more comprehensible to a broader range of readers~\cite{al2021automated} by reducing language complexity in terms of vocabulary and sentence structure without distorting its original intent and meaning~\cite{sheang2021controllable}.


Estimating a text's readability is subjective and reader-dependent. For example, substituting complex terms could be considered a good solution for non-native speakers, while shortening and rephrasing the text may be more suitable for children who have limited literacy skills~\cite{agrawal2023control}. 

SimpleText\footnote{\url{http://simpletext-project.com/2023/clef/}}~\cite{overview23}, now in its third iteration as a part of CLEF ~\cite{ermakova2022automatic, overview21}, intends to advance simplification of scientific text for non-experts. The lab consists of three tasks: selecting passages to include in a simplified summary (task 1), identifying and explaining difficult concepts for general audiences (task 2), and scientific text simplification (task 3)~\cite{overview23}.

This work exclusively focuses on task 3, which has already been tackled in previous iterations of SimpleText and beyond~\cite{monteiro2022using, rubio2022hulat, DBLP:journals/csl/HaslerGSWB17}. 
Recent advancements in language modeling, particularly transformer-based architectures like T5~\cite{t5} and different GPT variants~\cite{DBLP:journals/corr/abs-2005-14165, radford2018improving, radford2019language}, have enabled text simplification methods to incorporate semantic features in addition to lexical and syntactic ones. Notable works~\cite{wu2022cyut, talec2022using, monteiro2022using} primarily focus on using T5 models, while \citet{rubio2022hulat} leverage BART~\cite{BART}. Additionally, a GPT-2-based text simplification model was proposed in a zero-shot manner~\cite{mostert2022university}. 
Recently, ChatGPT~\cite{chatgpt} has emerged as a powerful tool that can be used without training. 

In one of our approaches for text simplification for non-experts, we use prompt engineering and complex phrase identification preprocessing with ChatGPT. Further, we contribute approaches based on the T5 and PEGASUS summarization models. 

\section{Dataset}\label{sec:dataset}


The dataset for task 3 of the SimpleText consists of short texts taken from scientific publications, primarily single sentences, e.g., \textit{"The malicious alteration of machine time is a big challenge in computer forensics."} (sentence ID~\texttt{G09.1\_567125047\_1}). The SimpleText lab offers a train and a test dataset split. The test split is categorized into small, medium, and large datasets, with the texts from the small being included in the medium and the large split containing all texts from the other two. In our work, we utilized the training set to engineer prompts and implemented our approaches on the large dataset comprising 152.072 source texts. We encountered several issues with the large dataset. The dataset includes duplicated samples. Removing them reduces the number of unique texts in the dataset to 135.540. Some texts (e.g., all sentences of the document with the doc\_ID \texttt{2787296320}) appear up to five times in the dataset. 
These duplicate entries are identical, with a few exceptions where the query texts differ. Additionally, the dataset frequently contains utf-8 hexcodes within the texts, as shown in \autoref{tab:examples_monkey}. The next issue concerns samples where some texts are incomplete, as highlighted in the example mentioned in \autoref{sec:chatGPT_intro}. These issues pose challenges for the summarization models described in \autoref{sec:summarization_models}.
Further, a small number of texts are either empty or contain punctuation marks. This can result in AI hallucination or generating empty simplified texts, as illustrated in \autoref{tab:eval_measures}. The phenomenon of hallucination is an already recognized problem of large language models~\cite{zhang2023language}.

\section{Methodological Approaches to Simplification}
This section outlines the methodology employed in each of our submitted runs. 
We generated four runs, including one using ChatGPT prompting and three based on out-of-the-box text summarization models.

\subsection{ChatGPT for Simplification}\label{sec:chatGPT_intro}
In our \texttt{irgc\_task\_3\_ChatGPT\_2stepTurbo} run (called \texttt{ChatGPT} in the remainder of this paper), we employed a combination of complex phrase identification and simplification with ChatGPT. 

\begin{figure}[t]
    \centering
    \includegraphics[width=\textwidth]{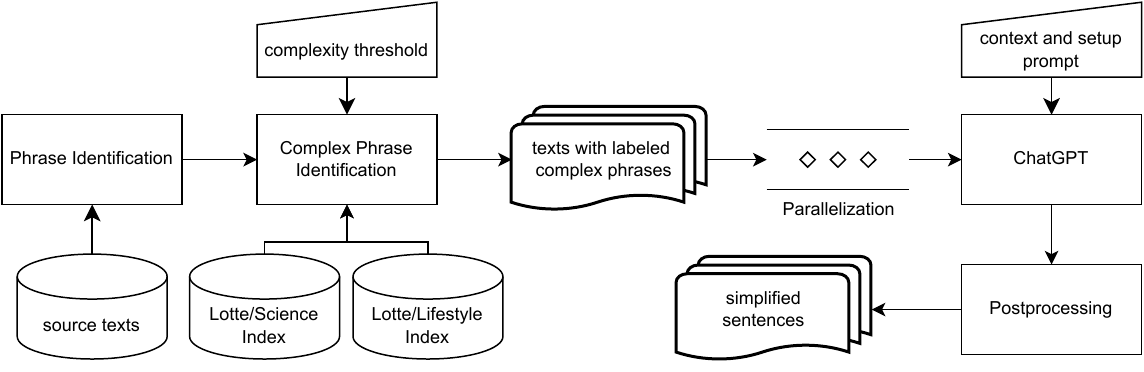}
    \caption{Pipeline of our \texttt{ChatGPT} run including the complex phrase identification component.}
    \label{fig:gpt_pipeline}
\end{figure}

The architecture of our two-step simplification approach using ChatGPT conjoined with complex phrase identification is depicted in \autoref{fig:gpt_pipeline}. 
In brief, we extract keyphrases from the source text before identifying complex phrases. 
The complex phrases are then labeled in the source text, which is given to ChatGPT with a prompt asking it to simplify the sentences with special focus on the labeled parts.

In our approach, we avoided performing any preprocessing.
However, we conducted some initial testing to gain better insights into the samples and the functionality of ChatGPT.
We witnessed that our model sometimes generated information not present in the source texts. 
We encountered this situation primarily when the sentences in the source text were sparse and a query text was provided in the prompt. 
For example, by including the query text \textit{"digital assistant"} along with the text \textit{"This approach is ma..."} (sentence ID~\texttt{G01.1\_3006661050\_2}), we noted that the text was rephrased as \textit{"This way of doing things is commonly used in digital assistants."}. 
To address this issue, we decided not to include the provided query texts in our approach. 
When we tried to simplify multiple source texts at once using ChatGPT, the model inferred information from consecutive sentences despite the prompt telling the model to treat these texts separately. 
Consequently, we randomized the order of the source texts.

For our submitted run, the first step of the approach consists of phrase identification and complex phrase identification. 
These labeled texts are then provided to the ChatGPT module as a second step for simplification. 
Finally, the simplified texts are post-processed. 
Since the complex phrase identification, simplification, and post-processing steps are fundamental components of our approach, we will elaborate on them in detail.

\paragraph{Complex Phrase Identification}\label{sec:gpt_other_components}
One aspect of our simplification approach is to avoid highly complex phrases, i.e., phrases that are difficult to understand for non-experts.
We first identify keyphrases from the source text to make this possible. Keyphrases are those phrases of a text which contain core information~\cite{kulkarni2022learning}.
We assume that especially keyphrases are relevant for understanding texts, so we try to express this core information more easily.
Since identifying keyphrases is an already established task, we use \textit{KBIR-inspec}, a pre-trained model which is available on Hugginface\footnote{\url{https://huggingface.co/ml6team/keyphrase-extraction-kbir-inspec}}.

However, not all recognized phrases are complex, so we introduce a mechanism to distinguish complex from non-complex phrases.
To do this, we evaluate the complexity of all terms in a phrase (\autoref{complex_f}) and whether the terms' aggregated complexity value exceeds a manually specified threshold.
Since the domain of the task is scientific texts, we evaluate the complexity of terms using the term statistics of two text corpora: a dataset that is comprised of texts from lifestyle forums and one that is comprised of texts from science-focused forums~\cite{Santhanam2021ColBERTv2}. 
We assume that terms whose relative frequency in scientific texts is greater than those from lifestyle texts also tend to be more complex.
We obtain our complexity estimate by the difference of the inverse document frequency (\textit{idf}) of term $t$ from the lifestyle dataset with the \textit{idf} value of the scientific dataset. The complexity of a phrase $<t_1, ..., t_n>$ is defined by the function $\phi$:
\begin{equation}
\label{complex_f}
    \phi(<t_1, ..., t_n>) = \max_{i \in \{1,..,n\}} \left\{\log_{N_{lf}}\left(\frac{N_{lf}}{df_{lf}(t_i)} \right) - \log_{N_{sc}}\left(\frac{N_{sc}}{df_{sc}(t_i)}\right)  \right\}.
\end{equation}

Here $df_{lf}(t)$ is the number of documents from the lifestyle dataset in which the term $t$ appears, $df_{sc}(t)$ for the science dataset respectively.
The total number of documents is $N$. We set the complexity threshold to 0.01. So every phrase above this threshold gets tagged as complex. In our case, tagging means annotating phrases using square brackets (e.g., second row in \autoref{tab:examples_noaron}). 
Thus, the more frequently the terms of a phrase occur in scientific texts, and the less frequently they occur in lifestyle texts, the more complex we estimate the phrase to be. To index and analyze the datasets and term statistics, we use the PyTerrier~\cite{pyterrier2020ictir} framework. 

\paragraph{ChatGPT module}\label{sec:chatGPT_process}
The ChatGPT simplification module involves simplifying the sentences, with a focus on the labeled complex phrases. In this work, we employ the \textit{gpt-3.5-turbo-0301} model accessed via API\footnote{\url{https://platform.openai.com/docs/models/gpt-3-5}}.

The model is given a first instruction prompt\footnote{
    The used prompt is: \textit{"You are given 10 texts, TREAT THEM SEPARATELY. 
Complex technical or scientific terms are indicated by square brackets (e.g. [convolutional neural network]).
For each text, replace or explain the terms in square brackets to make it understandable to non-experts and to simplify for non-expert readers. 
Make sure that you return exactly one enumerated list of 10 simplified texts and that all terms in square brackets are simplified or explained!"}
}.
The prompt was iteratively engineered by testing, with the goal of circumventing common problems encountered with ChatGPT.
In addition to a general description of the task to simplify a list of input texts, the prompt contains further specifications.
The prompt includes an instruction to treat a list of input terms separately to prevent information from diffusing between sentences.
We further provide information on the number of expected input texts.
This asserts that the output will not split input texts with more than one sentence.
The essential part of the initial prompt is explaining how complex phrases are marked by square brackets, along with an example. 
The system is requested to either explain or replace these complex phrases.
The list of ten input sentences with the complex phrases tagged by the complex phrase identification module is provided to the system as a separate user-tagged message as a Markdown-formatted list.
The marking of complex phrases by square brackets seems to work well with ChatGPT.

\paragraph{Postprocessing}
As a first postprocessing step, we remove list tags and split the text generated by ChatGPT into separate simplified texts. 
This is necessary because ChatGPT does not reliably return a cohesively structured list of results despite the same prompt structure being used.
The results were in all cases formatted as a list of texts, but sometimes the lists were enumerated, while some contained dashed or other itemization marks.
Furthermore, ChatGPT sometimes separated the list elements by empty lines. 
Despite the detailed description in the prompt, ChatGPT sometimes did not remove terms in square brackets, and in some of these instances, the square brackets were not removed either.
We apply a postprocessing step to cope with these cases, removing all square brackets from all simplified texts.

\subsection{Summarization Models for Simplification}\label{sec:summarization_models}

In addition to the \texttt{ChatGPT} run, we also conducted three additional runs using out-of-the-box sequence-to-sequence models popular for summarization tasks.
The main reason for employing text summarization techniques is to attempt to simplify the text snippets. 
In this scenario, we have two runs based on the T5~\cite{t5} model and one run that uses PEGASUS~\cite{pegasus}.
For both models we have the exact same processing steps: We instruct the model to \textit{summarize} all source texts. We do not consider the queries but only the source sentences, so the \textit{source\_snt}. We use batch processing of texts which we want to simplify, set the maximum input length to 512 tokens per example, pad the inputs, and determine outputs to have a maximum length of 100 tokens. 
This leads to a model occasionally creating multiple sentences in one summarization, each fully summarizing the original text. Therefore, out of these merged simplifications (summarizations), we pick the shortest sentence as the simplified version of the original text.

In our \texttt{irgc\_task\_3\_t5} run (called \texttt{t5} for simplicity) we use the T5~\cite{t5} base model with 220 million parameters available on Huggingface~\footnote{\url{https://huggingface.co/t5-base}}. 

For our \texttt{irgc\_task\_3\_t5\_noaron} run (called \texttt{t5\_noaron} for simplicity), we use the output from the \texttt{t5} run but exclude a reoccurring hallucination. In the \texttt{t5} runs the simplified sentences contain the name of a specific healthcare researcher several times, indicating that person stated the following simplification of the source text, e.g., text ID~\texttt{G04.3\_2094757405\_4} \textit{"aaron carroll: modelling of an infection network between viral and cellular proteins will provide a conceptual framework"}. In the \texttt{t5\_noaron} run we remove this hallucination by removing the expression \textit{"aaron carroll: "} from all texts produced by the \texttt{t5} run.

In our \texttt{irgc\_task\_3\_pegasusTuner007plus\_plus} (called \texttt{pegasus} in the remainder of this paper for simplicity) run we use the PEGASUS~\cite{pegasus} summarization model, which is provided by Tuner007\footnote{\url{https://huggingface.co/tuner007/pegasus\_summarizer/discussions}} as a fine-tuned version of the original PEGASUS.

\section{Implementation and Generated Runs}\label{implementation_runs}

\subsection{Implementation Specifications}
We implemented all approaches in Google Colab\footnote{ChatGPT: \url{https://colab.research.google.com/drive/10LyozPzxUlqFxHkXyfjxezO469c1ou9z?usp=sharing}\\Summarization Models: \url{https://colab.research.google.com/drive/1dI0rGH2mPMJ8OdsGnrqAz0HWQzfQCxK0?usp=sharing}}.
The \texttt{pegasus} and \texttt{t5} runs were generated on an A100 GPU. 
Rephrasing the sentences using T5 took approximately 90 minutes, using PEGASUS took approximately 4.5 hours.

In order to reduce both the cost of ChatGPT calls and the duration of execution, we took the following approaches.
A call to ChatGPT was designed to encapsulate a prompt containing ten sentences as one request to reduce overhead. 
In addition, we split the dataset into 15 chunks to better handle unexpected problems.
To reduce execution time, we processed each chunk using 75 processes. This allowed us to reduce the expected total computation time from 60 to 20 hours.
We delayed the start of each process because the ChatGPT API has a request limit of 3,500 requests/minute and 90,000 tokens/minute. Therefore, exceeding the rate limit yields an unusable result for the request. 
According to our estimation, processing ten texts takes about 14 seconds (ten seconds for pre- and post-processing and four seconds for the ChatGPT API call).
The ChatGPT API incurred a cost of about \$30 for the entirety of our runs.

\subsection{Results}
In order to thoroughly analyze the outputs of our runs, we conducted an \textit{automatic} and an example-based \textit{manual} evaluation of generated simplifications in comparison to the source texts. These evaluations lead to our final estimation of \textit{priorities} for the produced runs. The priorities indicate our estimation of the quality of the produced results and which runs' results should be considered in SimpleText's manual evaluation.\footnote{We assumed, that a manual evaluation would be part of the evaluation of the submissions, which is not the case. However, we believe manual evaluations as described in our manual evaluation in section \ref{sec:man_eval} better reflect the quality of the produced texts than the metrics used in the SimpleText evaluation.}

\subsubsection{Automatic Evaluation}

\begin{table}[t]
    \centering
    \footnotesize
    \begin{tabular}{l|l|l|l|l|l}
         measure & \texttt{source texts} & \texttt{ChatGPT} & \texttt{t5} & \texttt{t5\_noaron} & \texttt{pegasus}  \\ \hline
    empty texts $\downarrow$ & - & 3 & 148 & 148 & \textbf{1}\\

compression $\uparrow$	& -	&	1.16 $\pm$ 0.72	&	1.53 $\pm$ 2.3	&	\textbf{1.55} $\pm$ 2.34	&	1.0 $\pm$ 1.6	\\

Flesch $\uparrow$	&	36.7 $\pm$ 23.33	&	42.41 $\pm$ 21.61	&	\textbf{46.4} $\pm$ 38.65	&	46.3 $\pm$ 38.85	&	38.79 $\pm$ 23.28	\\
Dale-Chall $\downarrow$	&	12.68 $\pm$ 2.18	&	12.35 $\pm$ 1.96	&	12.41 $\pm$ 3.05	&	12.37 $\pm$ 3.05	&	\textbf{11.44} $\pm$ 2.62	\\
difficult words $\downarrow$	&	8.51 $\pm$ 4.32	&	7.55 $\pm$ 3.33	&	6.23 $\pm$ 2.86	&	\textbf{6.2} $\pm$ 2.88	&	7.99 $\pm$ 3.83	\\
reading time $\downarrow$	&	1.94 $\pm$ 0.99	&	1.71 $\pm$ 0.68	&	1.45 $\pm$ 0.59	&	\textbf{1.44} $\pm$ 0.59	&	2.24 $\pm$ 1.03	\\
syllable count $\downarrow$	&	39.12 $\pm$ 20.07	&	34.4 $\pm$ 13.81	&	29.22 $\pm$ 12.22	&	\textbf{29.11} $\pm$ 12.25	&	45.23 $\pm$ 20.86	\\
lexicon	count $\downarrow$&	22.51 $\pm$ 11.05	&	20.26 $\pm$ 8.04	&	16.88 $\pm$ 7.04	&	\textbf{16.8} $\pm$ 7.07	&	26.18 $\pm$ 11.73	\\
sentence count	$\downarrow$ &	\textbf{1.05} $\pm$ 0.58	&	1.07 $\pm$ 0.29	&	1.44 $\pm$ 0.56	&	1.44 $\pm$ 0.56	&	1.32 $\pm$ 0.62	\\
         
    \end{tabular}
    \caption{Averaged readability scores and their standard deviations for source texts as well as our four run variants. $\downarrow$ indicates the measure is better the lower it is, $\uparrow$ indicates the measure is better the higher it is.}
    \label{tab:eval_measures}
\end{table}

In our automatic evaluation, we primarily focused on assessing the readability of the generated text rather than other metrics such as semantic accuracy. We believe that in the context of text simplification, readability measures play a crucial role in evaluating the quality of the output.

We observed readability measures that only require text snippets and no gold-labeled simplifications.
Table~\ref{tab:eval_measures} indicates the following averaged readability measures for the source texts as well as our four runs on the large dataset.
The number of empty texts indicates how many of the texts a run produced did not contain any content. The following measures have all been averaged by the number of texts which are not empty.
The compression ratio indicates by how much the original source text's size has been reduced in the simplified variant.
The following readability measures have all been implemented using the Python library textstat\footnote{\url{https://pypi.org/project/textstat/}}:
The Flesch readability index~\cite{Flesch1948ANR} quantifies the reading ease of a text based on word and sentence length. 
Scores between 30 and 49 indicate difficult text.
The new Dale-Chall~\cite{chall1995readability} score indicates the reading level of a text as a grade corresponding to the familiarity of persons from that grade with a list of the 3000 most common English words. 
Scores are defined up to a value between 9 and 9.9 which corresponds to the reading level of an average college student. The higher the score, the more difficult a text.
The number of difficult words indicates the number of words with $\ge$ 3 syllables and which are not in a predefined list of easy words.
The reading time indicates the seconds required to read a text; each character taking 14.69ms~\cite{reading_time}. 
The syllable count gives the number of syllables in a text.
The lexicon count gives the number of different words in a text.
The sentence count indicates how many sentences a text consists of. For these last three measures, one could argue that a simplified text should be shorter in terms of containing fewer syllables, words and sentences compared to the original text which also is in line with previous work~\cite{martin2019controllable}. The artificially constructed example in \autoref{tab:counts} shows that this oversimplified assumption does not always hold.

\begin{table}[]
    \centering
    \begin{tabularx}{\textwidth}{l|X|l|l|l}
         type & text & syllables & lexicons & sentences\\\hline
         source & The WHO reprimands the potentate for using UVAs. &13&8&1\\\hline
         manually simplified &The world health organization spoke a warning. The ruler should not be using unmanned areal vehicles (drones).&27&17&2 \\
    \end{tabularx}
    \caption{Artificially constructed example which has not been taken from the dataset showcasing the non-trivial relation between the length of a simplified text and its source text.}
    \label{tab:counts}
\end{table}

From our strictly numerical analysis of the runs, we found that all our runs improved the source texts' readability in all measures except for the number of sentences. The \texttt{t5\_noaron} run yielded among the best results for many of the observed measures. The \texttt{ChatGPT} run did not produce the numerically best results. It seemed to produce texts with more syllables and lexicals than the T5 models which negatively impacted the reading time as well. 

As a result of our automatic evaluation of readability measures which do only depend on text snippets and not on some form of gold labels, we would expect the \texttt{t5\_noaron} run to provide the best results.

\subsubsection{Manual Evaluation}
\label{sec:man_eval}

To support our decision-making process on the estimation of the quality of runs we conducted an example-based manual evaluation.
We randomly selected texts to assess perceived readability, grammar, and vocabulary and how the simplified texts cope with complex scientific terms. This evaluation process was similar to the human evaluation described by~\citet{ermakova2022automatic}.

\autoref{tab:examples_noaron} shows an example for a source text along with the tagged text produced by the complex phrase identification module of the ChatGPT approach as well as the texts of the runs.

Our manual evaluation yielded several interesting observations. The results did not align with the findings of our automatic evaluation. 
Furthermore, the PEGASUS model produces texts, that are almost identical to the original text, which has been described as a common problem in last year's iteration of SimpleText~\cite{DBLP:conf/ecir/ErmakovaSHAAK23}.
Another finding is that the text produced by the T5 model is significantly shorter and grammatically simpler but omits important information from the source sentence, e.g., the impact on drug discovery. 
Further, the text still contains scientific formulations (\textit{"conceptual framework"}) and a case of AI hallucination in the form of reporting the text as a quote, which has been partially filtered out in the \texttt{t5\_noaron} approach.

From our perspective, the ChatGPT approach yielded the best results by far.
The identification of phrases that need to be replaced or explained for a non-expert reader was successful.
The ChatGPT model replaced or explained most of the phrases with some exceptions such as \textit{"infection network"}, which sub-optimally was replaced with \textit{"infection pathways"}.
Overall, however, the text produced by the model conveys all essential information from the source sentence in a simpler, concise way.

Another factor influencing our decision to rank the ChatGPT approach as best performing in the manual evaluation is the ability of the system to cope with the problems in the dataset discussed in \autoref{sec:dataset} as well as correctly interpreting abbreviations and scientific terms without further context. 
The example shown in \autoref{tab:examples_monkey} illustrates these advantages, showing how the model correctly interprets the utf-8 fragments (\textit{"NHPu0027s"}) as well as interpreting \textit{"NHP"} as \textit{"non-human primates"}.


 \begin{table}[t]
    \centering
    \small
    \begin{tabularx}{\textwidth}{l|X}
    
         source text & Indeed, modelling of an infection network between viral and cellular proteins will provide a conceptual and analytic framework to efficiently formulate new biological hypothesis at the proteome scale and to rationalize drug discovery.\\\hline
         tagged text & Indeed, modelling of an [infection network] between viral and cellular proteins will provide a conceptual and analytic framework to efficiently formulate new [biological hypothesis] at the [proteome] scale and to rationalize [drug discovery].\\\hline
         \texttt{ChatGPT} & Building a model of infection pathways between viral and cellular components can help scientists generate new ideas for drugs at a large scale.\\\hline
         \texttt{t5} & aaron carroll: modelling of an infection network between viral and cellular proteins will provide a conceptual framework.\\\hline
         \texttt{t5\_noaron} & modelling of an infection network between viral and cellular proteins will provide a conceptual framework.\\\hline
         \texttt{pegasus} & Modelling of an infection network between viral and cellular proteins will provide a conceptual and analytic framework to efficiently formulate new biological hypothesis at the proteome scale and to rationalize drug discovery.
     \end{tabularx}
    \caption{Examples for the output of our runs, including the complex term identification module of the ChatGPT approach, for an exemplary source sentence (ID~\texttt{G04.3\_2094757405\_4}). In the manual evaluation, we found the ChatGPT run to produce the best simplifications for a non-expert audience.}
    \label{tab:examples_noaron}
\end{table}

\begin{table}[t]
    \centering
    \small
    \begin{tabularx}{\textwidth}{X|X}
    source text & \texttt{ChatGPT}\\\hline
    There are two challenges in analyzing the NHPu0027s surveillance video: the NHPu0027s behaviors are lack of regularity and intention, and serious occlusions are brought by the fences of the cages. &
    The irregular behavior of non-human primates and occlusions caused by cages' fences make it challenging to analyze their surveillance video footage.
    \end{tabularx}
    \caption{Example for a source text with problematic formatting issues (ID~\texttt{G03.2\_2395758311\_4}). The ChatGPT approach did not only effectively deal with this issue but correctly interpreted the meaning of \textit{"NHP"} as \textit{"non-human primates"} from the given context of the source text.}
    \label{tab:examples_monkey}
\end{table}
\subsubsection{Priorities of Runs}
Due to the task description asking to simplify texts for non-experts, we weigh the overall soundness, completeness, and textual quality evaluated in the manual evaluation more than the results of the automatic evaluation.
This decision is consolidated by the fact that the priority determines which of the runs will be manually evaluated.
Further, the metrics for the automatic evaluation mostly reflect the readability for those with reduced reading capability, such as children, the most common application of text simplification. 
However, non-experts should be no less capable of reading texts with, e.g., longer words or generally higher complexity, rather the overall scientific specificity needs to be reduced. 
Accordingly, we give the \texttt{ChatGPT} run the highest priority, followed by the \texttt{t5\_noaron}, the \texttt{pegasus}, and the \texttt{t5} runs in that order.

\section{Conclusion}\label{conclusion}

Our research focuses on text simplification, particularly in the context of scientific texts for non-experts. We present one run employing ChatGPT with complex phrase identification and three runs using out-of-the-box summarization models (two based on T5 and one based on PEGASUS).

While our automatic evaluation did not rank \texttt{ChatGPT} as the best run, a manual analysis evaluated the texts produced through \texttt{ChatGPT} as the best. Although we did not explicitly evaluate the inclusion of complex phrase identification in the \texttt{ChatGPT} run, we found it to improve the system's effectiveness. The identified complex terms indicate that the datasets for constructing the complex phrase identification system were a reasonable choice.

During our implementation, we encountered several issues. Labeling complex phrases using square brackets in our approach may pose challenges when the input text already contains square brackets. This problem could be circumvented by implementing an additional preprocessing step. Another challenge is hallucinated content, possibly attributed to the absence of context in data.
A possible improvement for the summarization model approaches would be flagging difficult words as ones we want to exclude in the simplified (summarized) variant.





\bibliography{ref.bib} 

\end{document}